\documentclass[letterpaper]{article} 
\usepackage{aaai2026}  
\usepackage{times}  
\usepackage{helvet}  
\usepackage{courier}  
\usepackage[hyphens]{url}  
\usepackage{graphicx} 
\usepackage{natbib}  
\usepackage{caption} 
\usepackage{algorithm}
\usepackage{algorithmic}
\usepackage{amsfonts}
\usepackage{amsmath}
\usepackage{booktabs}
\usepackage{multirow}
\usepackage{makecell}
\usepackage{booktabs}
\usepackage{tabularx}
\usepackage{array}   
\newcolumntype{Y}{>{\centering\arraybackslash}X}
\usepackage{subcaption}

\usepackage{times}
\usepackage{helvet}
\usepackage{courier}
\usepackage{xcolor}

\usepackage{newfloat}
\usepackage{listings}
\DeclareCaptionStyle{ruled}{labelfont=normalfont,labelsep=colon,strut=off} 
\floatstyle{ruled}
\newfloat{listing}{tb}{lst}{}
\floatname{listing}{Listing}
\title{FLUX-Makeup: High-Fidelity, Identity-Consistent, and Robust Makeup \\ Transfer via Diffusion Transformer}
\author{
    Jian Zhu\textsuperscript{\rm 1 2 }\thanks{Equal contribution. \textsuperscript{\rm 1}Nanjing University of Science and Technology \textsuperscript{\rm 2}360 AI Research \textsuperscript{\rm 3}Beijing University of Aeronautics and Astronautics.
    Corresponding author: Dawei Leng (lengdawei@360.cn), Yang Xu (xuyangth90@njust.edu.cn).},
    Shanyuan Liu\textsuperscript{\rm 2 * },
    Liuzhuozheng Li\textsuperscript{\rm 2},
    Yue Gong\textsuperscript{\rm 2 3},
    He Wang\textsuperscript{\rm 1}, \\
    Bo Cheng\textsuperscript{\rm 2},
    Yuhang Ma\textsuperscript{\rm 2},
    Liebucha Wu\textsuperscript{\rm 2},
    Xiaoyu Wu\textsuperscript{\rm 2},
    Dawei Leng\textsuperscript{\rm 2}, 
    Yuhui Yin\textsuperscript{\rm 2},
    Yang Xu\textsuperscript{\rm 1} 
}
\affiliations{
    
}

\usepackage{bibentry}

\begin{document}

\maketitle

\begin{abstract}
Makeup transfer aims to apply the makeup style from a reference face to a target face and has been increasingly adopted in practical applications. Existing GAN-based approaches typically rely on carefully designed loss functions to balance transfer quality and facial identity consistency, while diffusion-based methods often depend on additional face-control modules or algorithms to preserve identity. However, these auxiliary components tend to introduce extra errors, leading to suboptimal transfer results.
To overcome these limitations, we propose FLUX-Makeup, a high-fidelity, identity-consistent, and robust makeup transfer framework that eliminates the need for any auxiliary face-control components. Instead, our method directly leverages source–reference image pairs to achieve superior transfer performance. Specifically, we build our framework upon FLUX-Kontext, using the source image as its native conditional input. Furthermore, we introduce RefLoRAInjector, a lightweight makeup feature injector that decouples the reference pathway from the backbone, enabling efficient and comprehensive extraction of makeup-related information.
In parallel, we design a robust and scalable data generation pipeline to provide more accurate supervision during training. The paired makeup datasets produced by this pipeline significantly surpass the quality of all existing datasets. Extensive experiments demonstrate that FLUX-Makeup achieves state-of-the-art performance, exhibiting strong robustness across diverse scenarios. 

\end{abstract}


\section{Introduction}
With the rapid development of e-commerce and virtual try-on technologies, makeup transfer has emerged as a crucial task in social media, beauty applications, and digital content creation. The goal is to seamlessly apply the makeup style from a reference face image onto a source face image while preserving the source identity and faithfully reproducing the reference makeup. Benefiting from advances in computer vision and deep learning, recent approaches \cite{deng2021spatially, kips2020gan, wan2022facial, yan2023beautyrec, gu2019ladn, li2018beautygan, liu2021psgan++} have achieved remarkable progress and demonstrated impressive results in various scenarios. Nevertheless, delivering consistently high-fidelity transfer across a wide spectrum of complex and diverse makeup styles remains a significant challenge.

\begin{figure}[t]
    \centering
    \includegraphics[width=0.47\textwidth]{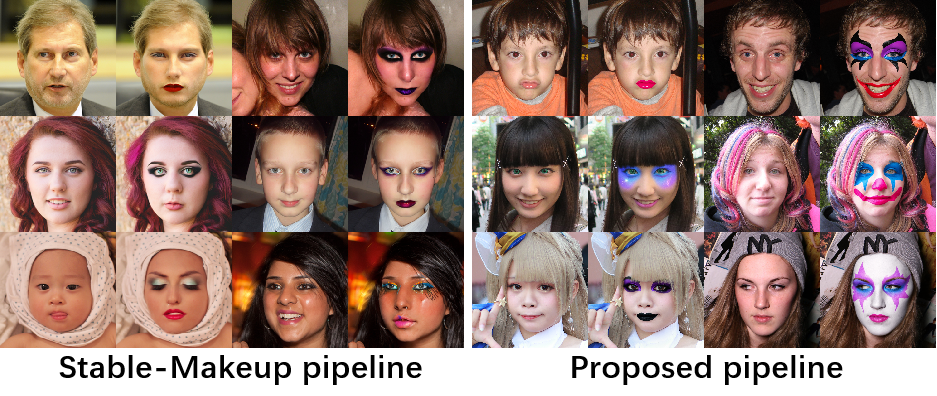}
    \caption{Comparison between the paired makeup data generated by our proposed pipeline and that produced by Stable-Makeup, showing that our data exhibits an overwhelming advantage in both consistency and makeup diversity.}
    \label{fig0}
\end{figure}

Due to the high cost of annotating paired makeup datasets and the limited scalability of predefined styles, most existing makeup transfer methods adopt unpaired training. While this paradigm avoids the need for expensive annotations, it suffers from significant facial misalignment, making stable and accurate supervision challenging. GAN-based \cite{goodfellow2020generative} approaches typically compensate for this by incorporating spatial priors or facial alignment techniques (e.g., landmark detection), but these additional cues increase model complexity and often introduce extra errors. Moreover, the lack of well-aligned pairs forces models to rely on weak surrogate losses, such as color histogram matching \cite{li2018beautygan, nguyen2021lipstick} or earth mover’s distance, which compromise robustness and fidelity.

Diffusion-based methods have recently shown strong generative capabilities in makeup transfer. However, many existing approaches are typically trained with an L2 reconstruction loss, which tends to rely on paired data to ensure stable supervision. To overcome the scarcity of real paired makeup datasets, some works attempt to synthesize pseudo-pairs. For example, Stable-Makeup \cite{zhang2025stablemakeup} generates training pairs using an image editing algorithm, while SHMT \cite{sun2024shmt} only takes reference images and extracts identity information via a 3D face model \cite{guo2020towards} combined with Laplacian-based contour extraction, effectively creating a form of synthetic paired data. Despite these efforts, the quality of the generated pairs remains low, exhibiting poor consistency and inaccurate alignment. Consequently, models trained on such data often rely on additional face-control modules \cite{zhang2023adding} to preserve identity during transfer, but these auxiliary components tend to introduce extra errors and frequently lead to suboptimal transfer results.

To address these limitations, we propose FLUX-Makeup, a novel diffusion-based framework that enables high-fidelity, identity-consistent, and robust makeup transfer without relying on any auxiliary face-control modules. Unlike previous methods, FLUX-Makeup directly leverages source–reference image pairs as its native conditional input, avoiding unnecessary intermediate cues and minimizing error propagation.
Specifically, our framework is built upon FLUX-Kontext \cite{flux2024, labs2025flux1kontextflowmatching}, where the source image serves as its inherent conditional input. For the reference image, we introduce RefLoRAInjector, a lightweight makeup feature embedding module that decouples the reference pathway from the backbone, enabling efficient and comprehensive extraction of makeup characteristics.  In parallel, we design a robust paired data generation pipeline that produces diverse and high-quality makeup pairs, overcoming the limitations of existing synthetic data approaches and providing accurate supervision for our model. Fig.~\ref{fig0} further illustrates the significant advantages of our generated data.
In summary, our main contributions are as follows:

\begin{itemize}
    \item We propose FLUX-Makeup, a high-fidelity, identity-consistent, and robust makeup transfer framework. It eliminates the need for any auxiliary face-control modules, enabling a straightforward and stable training paradigm. To the best of our knowledge, this is the first work that leverages DiT architecture for makeup transfer.
    \item We introduce RefLoRAInjector, a lightweight module designed for embedding makeup-specific information. It decouples the reference pathway from the backbone, significantly accelerating the transfer process and effectively guiding the injection of makeup features.
    \item We design a robust and scalable pipeline that produces diverse, well-aligned makeup pairs, substantially surpassing existing datasets in quality and consistency, and providing stronger supervision that significantly enhances model training and performance.
    \item Extensive experiments demonstrate that FLUX-Makeup not only achieves state-of-the-art performance but also exhibits strong robustness and generalization across diverse makeup styles and challenging scenarios.
\end{itemize}

\begin{figure*}[t]
    \centering
    \includegraphics[width=0.99\textwidth]{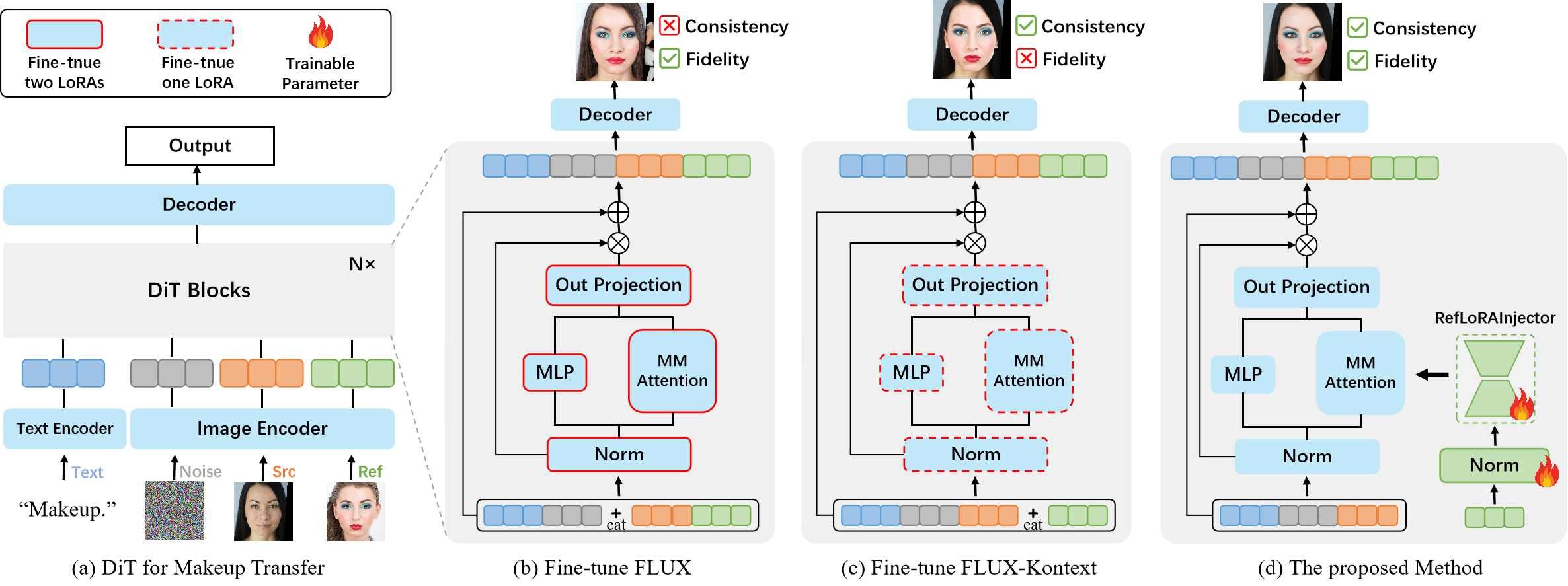}
    \caption{The overall framework of the proposed FLUX-Makeup. (a) A DiT-based architecture for makeup transfer. (b) A baseline method using the mainstream FLUX fine-tuning strategy, where source and reference images are concatenated in latent space and fine-tuned independently with two LoRA modules (followed by EasyControl). This achieves acceptable transfer performance but suffers from poor identity consistency. (c) A FLUX-Kontext-based fine-tuning method that conditions on the source image and appends the reference image afterward. It uses a single LoRA module and preserves background well, but the excessive consistency often results in direct copying of the reference face. (d) Our proposed method builds on FLUX-Kontext by introducing RefLoRAInjector, which effectively balances high-fidelity makeup transfer and identity preservation.}
    \label{fig1}
\end{figure*}

\section{Related Work}
\subsection{Makeup Transfer}
Makeup transfer has seen remarkable progress in recent years, early approaches were predominantly GAN-based, focusing on transferring styles between reference and source faces. BeautyGAN \cite{li2018beautygan} pioneered this direction by employing cycle-consistency for unpaired training, while BeautyGlow \cite{chen2019beautyglow} enhanced flexibility with disentangled latent representations, enabling smooth interpolation of makeup styles. PSGAN \cite{jiang2020psgan} adopted a patch-based strategy to better capture local makeup details, and CPM \cite{nguyen2021lipstick} leveraged correspondence learning to improve spatial alignment between source and reference faces.
Subsequent works further refined these techniques. SCGAN \cite{deng2021spatially}, RamGAN \cite{xiang2022ramgan}, and EleGANt \cite{yang2022elegant} incorporated attention mechanisms, region-aware transfer, and semantic decomposition to enhance identity preservation and style fidelity. Despite these advancements, GAN-based methods still face challenges, including the need for carefully crafted loss functions to balance style transfer and identity consistency, as well as heavy reliance on spatial priors or landmark detection for accurate facial alignment.

Recently, diffusion models have emerged as a powerful alternative due to their strong generative capabilities. Stable-Makeup \cite{zhang2025stablemakeup} leveraged diffusion models for makeup transfer by generating synthetic paired data through image editing algorithms, while SHMT \cite{sun2024shmt} adopted a 3D face model with Laplacian-based contour extraction to create pseudo-pairs. Although these approaches demonstrated promising results, their reliance on low-quality synthetic pairs limited supervision quality. To preserve identity, they further incorporated auxiliary face-control modules, but such components often introduced additional errors and degraded the final transfer quality.

\subsection{Diffusion models}
Diffusion models \cite{dhariwal2021diffusion} have emerged as a powerful class of generative models, achieving state-of-the-art performance across diverse image synthesis tasks, including text-to-image generation \cite{rombach2022high, podell2023sdxl}, image editing \cite{hou2024high, ma2024adapedit}, and subject-driven synthesis \cite{ruiz2023dreambooth, zhang2024ssr}. Their iterative denoising process enables creation of high-quality, diverse images, setting new benchmarks in image generation.
The seminal DDPM introduced a probabilistic framework that gradually refines noise into a coherent image through multiple denoising steps. Building upon this, DDIM \cite{song2020denoising} proposed a deterministic sampling strategy that accelerates inference while preserving image quality. To further cut computational costs and support high-resolution training, Latent Diffusion Models (LDMs) \cite{rombach2022high} were developed, leveraging a compressed latent space without compromising generation fidelity.

Subsequent advances have further expanded the capabilities of diffusion-based generation. Imagen \cite{saharia2022photorealistic} showcased the power of large-scale diffusion models in producing photorealistic images with strong text alignment, while Stable Diffusion \cite{rombach2022high} popularized latent diffusion by enabling efficient, open-source, high-resolution image synthesis. More recently, FLUX \cite{flux2024} introduced a transformer-based diffusion paradigm, combining the strengths of diffusion processes with the scalability of transformers. Building on this foundation, FLUX-Kontext \cite{labs2025flux1kontextflowmatching} further advanced conditional generation through contextual conditioning, offering enhanced control for complex generation tasks.

\section{Methodology}
\subsection{Architecture of FLUX-Makeup}
\paragraph{Makeup framework.} 
FLUX has demonstrated strong capabilities in image generation, making it an ideal foundation for our makeup transfer model. Specifically, we investigate two frameworks based on FLUX and FLUX-Kontext, respectively. In the FLUX-based framework, we adopt the common fine-tuning strategy used in DiT architectures like EasyControl \cite{zhang2025easycontrol} and OminiControl \cite{tan2024ominicontrol}. Here, the source and reference images are concatenated sequentially in the latent space, and LoRA \cite{hu2022lora} is applied to fine-tune the conditioning input layers. However, this approach often leads to conflicting objectives: the source image requires preserving identity, while the reference image aims to alter facial appearance. As illustrated in Fig.~\ref{fig1} (b), this conflict typically results in outputs with poor consistency between the background and face regions, as well as relatively low fidelity in makeup transfer.

In contrast, the FLUX-Kontext framework inherently incorporates a conditional image input, enabling improved preservation of both identity and background. In our implementation, the source image is directly used as the conditional input, while the reference image is concatenated in the latent space and fine-tuned via LoRA. Although this approach effectively enhances consistency, it can produce undesirable results—such as directly pasting the reference face onto the source image—due to overly strong conditioning on the input. This behavior is clearly visible in the output shown in Fig.~\ref{fig1} (c).

Motivated by these observations, we design our final model based on FLUX-Kontext to leverage its strong consistency preservation. To overcome the reference-pasting issue, we introduce RefLoRAInjector, an efficient module that injects makeup-related features by decoupling the reference image pathway from the backbone. This design effectively mitigates over-alignment, enabling accurate and robust makeup transfer.

\begin{figure*}[t]
    \centering
    \includegraphics[width=0.99\textwidth]{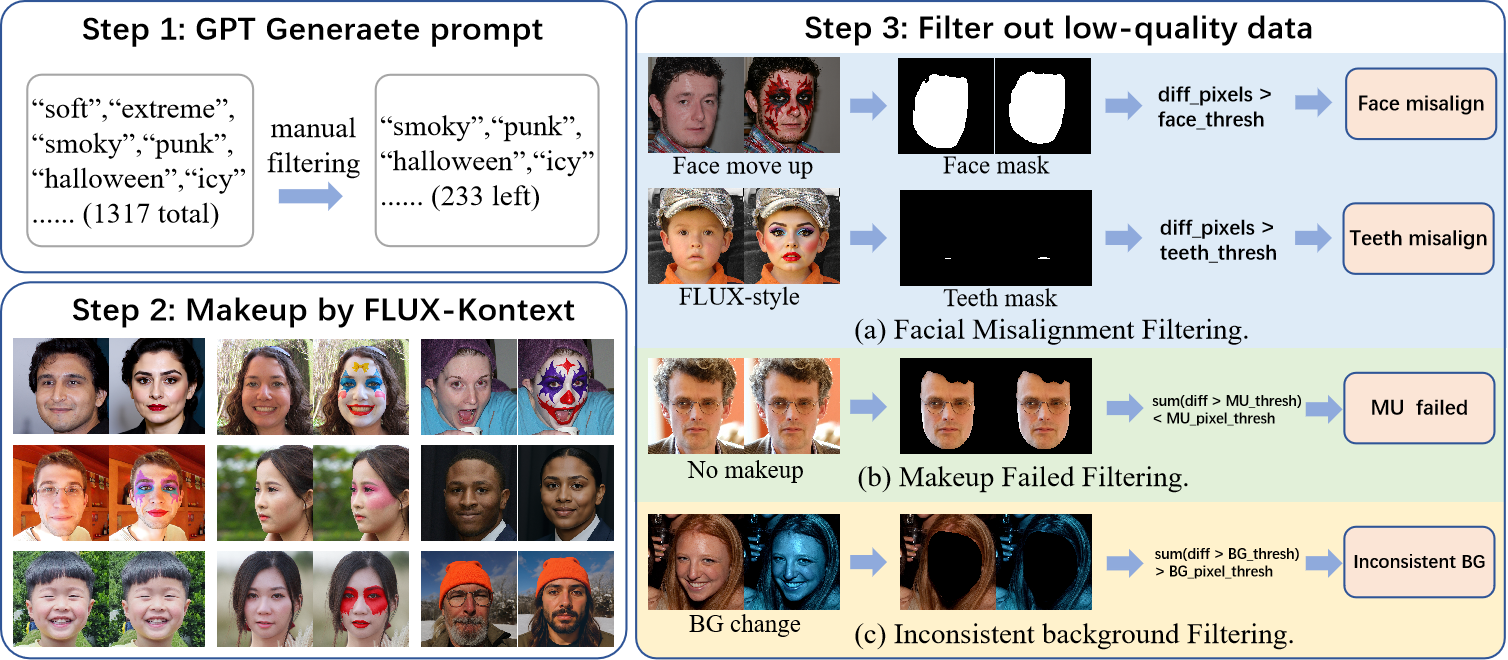}
    \caption{Pipeline for generating high-quality paired training data. (1) GPT is used to generate thousands of makeup-related keywords, which are then manually filtered based on the actual makeup effects produced by FLUX-Kontext; (2) The filtered prompts are applied to the FFHQ dataset using FLUX-Kontext to synthesize a large number of paired images; (3) Low-quality samples are removed by filtering from three perspectives: facial alignment, makeup success, and background consistency.}
    \label{fig2}
\end{figure*}

\paragraph{RefLoRAInjector.} 
To tackle the over-alignment issue in FLUX-Kontext—where the reference face is often directly pasted onto the source due to overly strong conditional consistency—we propose RefLoRAInjector, a lightweight and effective makeup feature injector that decouples reference information from the main generative pathway.
Specifically, RefLoRAInjector is implemented as a plug-in module within the self-attention layers. It takes reference features \(\mathbf{H}^{ref} \in \mathbb{R}^{B \times L \times d}\), obtained by encoding the reference image through the shared VAE encoder \cite{kingma2013auto}, and injects them into the attention key and value spaces via a low-rank adaptation mechanism.
Formally, two sets of low-rank projections are defined as:
\begin{equation}
\begin{aligned}
&\mathbf{K}^{ref} = W^{K}_{up}W^{K}_{down}\mathbf{H}^{ref}, \\
&\mathbf{V}^{ref} = W^{V}_{up}W^{V}_{down}\mathbf{H}^{ref},
\end{aligned}
\end{equation}
where \(W^{K}_{down}, W^{V}_{down} \in \mathbb{R}^{r \times d}, W^{K}_{up}, W^{V}_{up} \in \mathbb{R}^{d \times r}\), and \(r \ll d\) denotes the LoRA rank.
The injected reference projections \(\mathbf{K}^{ref}\) and \(\mathbf{V}^{ref}\) are then concatenated with the original key and value tensors along the sequence dimension:
\begin{equation}
\begin{aligned}
&\tilde{\mathbf{K}} = [\mathbf{K}, \mathbf{K}^{ref}], \\
&\tilde{\mathbf{V}} = [\mathbf{V}, \mathbf{V}^{ref}],
\end{aligned}
\end{equation}
where \([\cdot, \cdot]\) denotes concatenation. The enhanced key and value matrices \(\tilde{\mathbf{K}}\) and \(\tilde{\mathbf{V}}\) are then used for the subsequent attention computations.

By applying LoRA in a decoupled and targeted manner, RefLoRAInjector effectively regulates the influence of the reference image. This enables rich and expressive makeup styles to be transferred while avoiding identity collapse and background distortion. The module facilitates precise style conditioning, making it especially suitable for makeup transfer tasks that require a careful balance between source identity preservation and reference-driven editing.
As illustrated in Fig.~\ref{fig1} (d), the reference pathway is injected separately into the attention mechanism, ensuring modularity and extensibility. Compared to naive concatenation in the latent space, our approach achieves disentangled and controllable style modulation, resulting in enhanced robustness to diverse reference styles.

\begin{table*}[t]
\centering
\begin{tabularx}{\textwidth}{l|YYY|YYY|YYY} 
\toprule
\multirow{2}{*}{Methods} & \multicolumn{3}{c|}{MT} & \multicolumn{3}{c|}{Wild-MT} & \multicolumn{3}{c}{LADN} \\
\cmidrule{2-10}
& CLIP-I\(\uparrow\) & SSIM\(\uparrow\) & L2-M\(\downarrow\) & CLIP-I\(\uparrow\) & SSIM\(\uparrow\) & L2-M\(\downarrow\) & CLIP-I\(\uparrow\) & SSIM\(\uparrow\) & L2-M\(\downarrow\) \\
\midrule
CPM           &0.618&\underline{0.824}&\underline{9.15}  &0.565&\underline{0.810}&\underline{10.09} &0.642&0.811&11.25\\
PSGAN         &0.583&0.721&22.15 &0.556&0.538&28.04 &0.557&\underline{0.841}&12.62\\
EleGANt       &0.592&0.706&23.13 &0.575&0.522&29.24 &0.568&0.837&13.84\\
SSAT          &0.614&0.503&38.75 &0.583&0.500&36.50 &0.613&0.567&41.23\\
CSD-MT        &0.608&0.582&36.19 &0.572&0.474&36.48 &0.638&0.521&43.34\\
Stable-Makeup &\underline{0.649}&0.792&10.13 &\underline{0.661}&0.735&11.63 &\underline{0.715}&0.790&10.93\\
SHMT          &0.634&0.805&10.81 &0.645&0.793&10.29 &0.688&0.816&\underline{9.79}\\
FLUX-Makeup   &\textbf{0.668}&\textbf{0.879}&\textbf{5.18}  &\textbf{0.677}&\textbf{0.875}&\textbf{5.96}  &\textbf{0.731}&\textbf{0.862}&\textbf{5.20}\\
\bottomrule
\end{tabularx}
\caption{Quantitative results of CLIP-I, SSIM and L2-M on the MT, Wild-MT and LADN datasets.}
\label{compare}  
\end{table*}

\subsection{Data Generation Algorithm}
In this section, we present our approach for constructing high-quality paired makeup data. First, we leverage GPT4 \cite{achiam2023gpt} to generate over 1,000 makeup-related descriptive keywords (e.g., colorful, smoky). Each keyword is then evaluated by generating corresponding makeup images using the FLUX-Kontext model. Through manual inspection, we filter out unsuitable prompts—such as those prone to altering background information—and retain 233 high-quality makeup descriptors. Using this curated set, we employ FLUX-Kontext to synthesize paired makeup images based on the FFHQ dataset \cite{karras2019style}, resulting in an initial collection of paired data with diverse styles, as illustrated in Fig.~\ref{fig2}.

However, the raw dataset still contains numerous low-quality or invalid image pairs. After detailed analysis, we categorize the common failure cases into four types: (1) facial misalignment, (2) FLUX-style artifacts, (3) makeup failed, and (4) inconsistent background. To address these issues, we design and apply three targeted filtering strategies that effectively eliminate problematic pairs and substantially enhance the overall quality of the dataset.
\paragraph{Facial Misalignment Filtering.} 
To remove paired samples where facial structures become misaligned after makeup transfer, we employ a face parsing algorithm \cite{yu2018bisenet} to segment key facial regions, specifically the eyes, teeth, and facial contour. Other regions, such as the mouth, nose, and eyebrows, are excluded because they are often occluded by makeup, leading to unreliable segmentation results.
For each image pair, we extract binary masks for the selected regions before and after makeup application. Taking the face as an example, the masks are binary maps where 1 denotes the target region and 0 denotes the background. We then calculate the number of non-overlapping pixels between the two masks; if this value exceeds a predefined threshold \texttt{face\_thresh}, the pair is deemed misaligned and removed from the dataset. The effectiveness of this filtering strategy is illustrated in Fig.~\ref{fig2} (a).

The same procedure is applied to the eyes and teeth contour masks to ensure spatial consistency across all critical regions. Empirically, we also observe that severe FLUX-style artifacts—the second major failure mode—frequently co-occur with facial misalignment. Consequently, this filtering strategy effectively mitigates both issues simultaneously.
\paragraph{Makeup Failed Filtering.} 
In some cases, the FLUX-Kontext model fails to apply noticeable makeup to the face, even though the generated image remains well aligned with the source. These no-makeup pairs can pass the facial misalignment filter described earlier but still negatively impact training quality, making an additional filtering strategy essential.
To detect and remove such cases, we propose a pixel-level comparison method based on face region masking. First, the face parsing algorithm is used to obtain the full facial mask and extract the corresponding facial regions from both the source and generated images. We then compute the per-pixel absolute intensity difference between the two aligned regions. A pixel is considered modified by makeup if its absolute difference exceeds a predefined threshold \texttt{MU\_thresh}. By counting the number of modified pixels, we assess whether a meaningful makeup change has occurred. If the total count fails to exceed a second threshold \texttt{MU\_pixel\_thresh}, the sample is classified as a failed makeup case and is removed from the dataset.

\paragraph{Inconsistent Background Filtering.}
There are instances where the generated image preserves facial alignment and successfully transfers makeup, yet still presents background inconsistencies. These artifacts, though often subtle, lower the overall dataset quality and introduce noise that distracts the model from learning facial features effectively.
To handle this issue, we adopt a strategy similar to that used for detecting makeup failures. The face parsing algorithm is first applied to obtain the facial mask, and the background regions are isolated by excluding the face from each image. We then compute the absolute pixel-wise difference between the background regions of the source and generated images. A pixel is marked as inconsistent if its difference exceeds a predefined threshold \texttt{BG\_thresh}. If the total number of inconsistent pixels surpasses another threshold \texttt{BG\_pixel\_thresh}, the image pair is considered invalid and removed from the dataset.

\begin{figure*}[!t]
  \centering

  \begin{subfigure}{0.095\textwidth}
    \includegraphics[width=\linewidth]{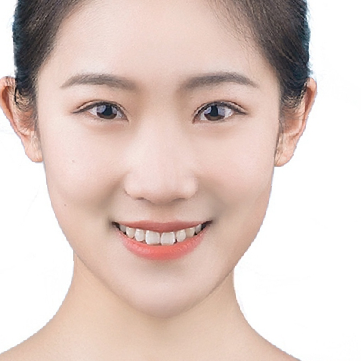}
  \end{subfigure}
  \hfill
  \begin{subfigure}{0.095\textwidth}
    \includegraphics[width=\linewidth]{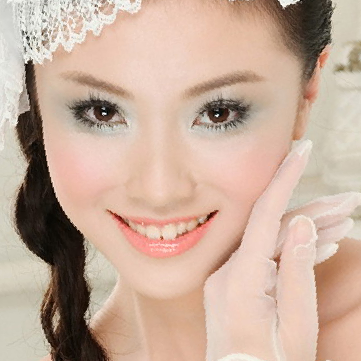}
  \end{subfigure}
  \hfill
  \begin{subfigure}{0.095\textwidth}
    \includegraphics[width=\linewidth]{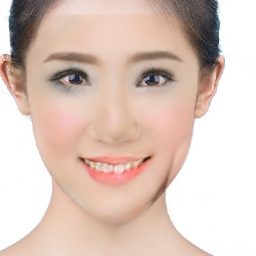}
  \end{subfigure}
  \hfill
  \begin{subfigure}{0.095\textwidth}
    \includegraphics[width=\linewidth]{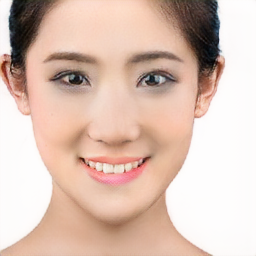}
  \end{subfigure}
  \hfill
  \begin{subfigure}{0.095\textwidth}
    \includegraphics[width=\linewidth]{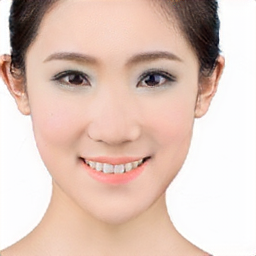}
  \end{subfigure}
  \hfill
  \begin{subfigure}{0.095\textwidth}
    \includegraphics[width=\linewidth]{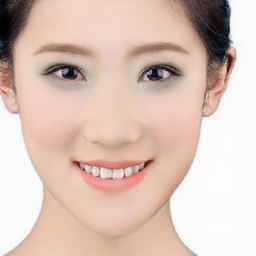}
  \end{subfigure}
  \hfill
  \begin{subfigure}{0.095\textwidth}
    \includegraphics[width=\linewidth]{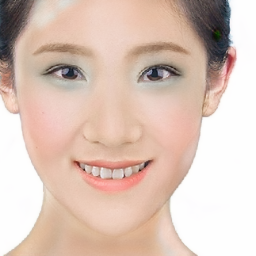}
  \end{subfigure}
  \hfill
  \begin{subfigure}{0.095\textwidth}
    \includegraphics[width=\linewidth]{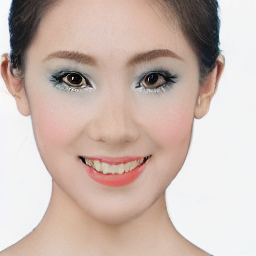}
  \end{subfigure}
  \hfill
  \begin{subfigure}{0.095\textwidth}
    \includegraphics[width=\linewidth]{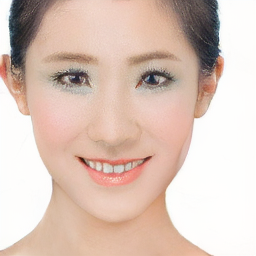}
  \end{subfigure}
  \hfill
  \begin{subfigure}{0.095\textwidth}
    \includegraphics[width=\linewidth]{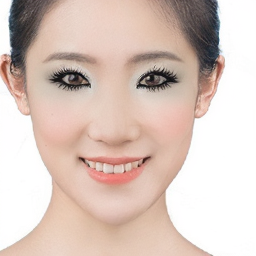}
  \end{subfigure}

  \vspace{0.2mm}

  \begin{subfigure}{0.095\textwidth}
    \includegraphics[width=\linewidth]{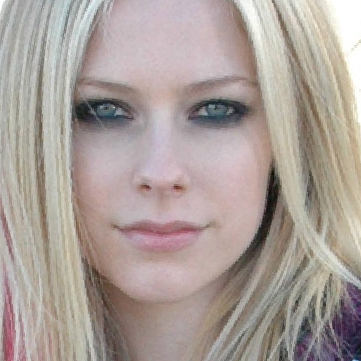}
  \end{subfigure}
  \hfill
  \begin{subfigure}{0.095\textwidth}
    \includegraphics[width=\linewidth]{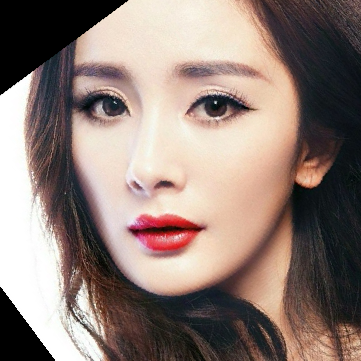}
  \end{subfigure}
  \hfill
  \begin{subfigure}{0.095\textwidth}
    \includegraphics[width=\linewidth]{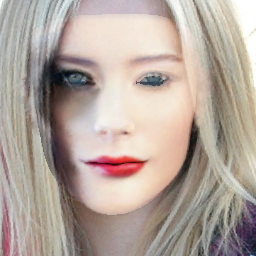}
  \end{subfigure}
  \hfill
  \begin{subfigure}{0.095\textwidth}
    \includegraphics[width=\linewidth]{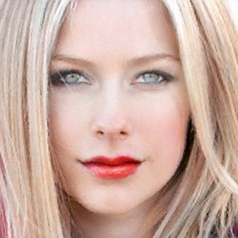}
  \end{subfigure}
  \hfill
  \begin{subfigure}{0.095\textwidth}
    \includegraphics[width=\linewidth]{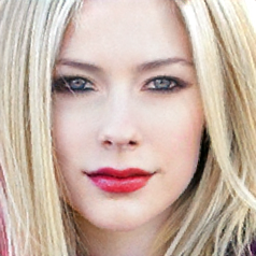}
  \end{subfigure}
  \hfill
  \begin{subfigure}{0.095\textwidth}
    \includegraphics[width=\linewidth]{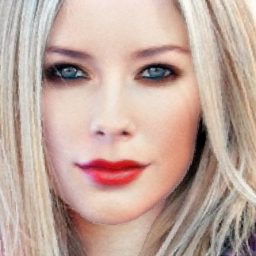}
  \end{subfigure}
  \hfill
  \begin{subfigure}{0.095\textwidth}
    \includegraphics[width=\linewidth]{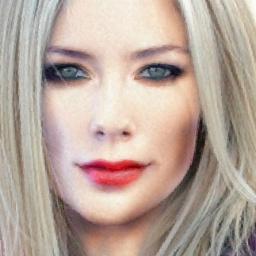}
  \end{subfigure}
  \hfill
  \begin{subfigure}{0.095\textwidth}
    \includegraphics[width=\linewidth]{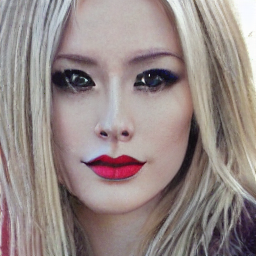}
  \end{subfigure}
  \hfill
  \begin{subfigure}{0.095\textwidth}
    \includegraphics[width=\linewidth]{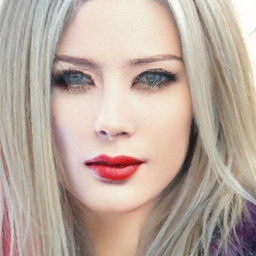}
  \end{subfigure}
  \hfill
  \begin{subfigure}{0.095\textwidth}
    \includegraphics[width=\linewidth]{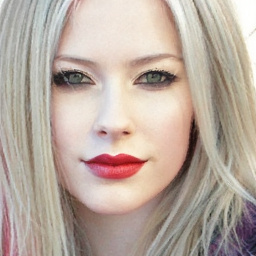}
  \end{subfigure}

  \vspace{0.2mm}

  \begin{subfigure}{0.095\textwidth}
    \includegraphics[width=\linewidth]{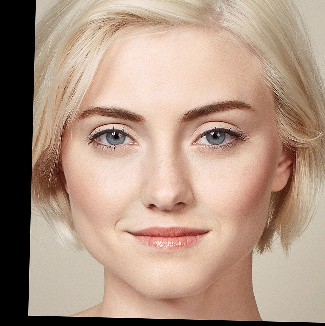}
  \end{subfigure}
  \hfill
  \begin{subfigure}{0.095\textwidth}
    \includegraphics[width=\linewidth]{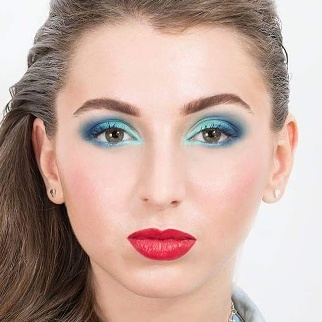}
  \end{subfigure}
  \hfill
  \begin{subfigure}{0.095\textwidth}
    \includegraphics[width=\linewidth]{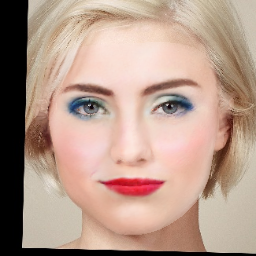}
  \end{subfigure}
  \hfill
  \begin{subfigure}{0.095\textwidth}
    \includegraphics[width=\linewidth]{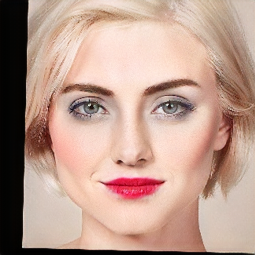}
  \end{subfigure}
  \hfill
  \begin{subfigure}{0.095\textwidth}
    \includegraphics[width=\linewidth]{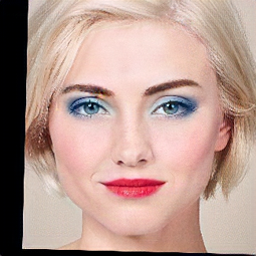}
  \end{subfigure}
  \hfill
  \begin{subfigure}{0.095\textwidth}
    \includegraphics[width=\linewidth]{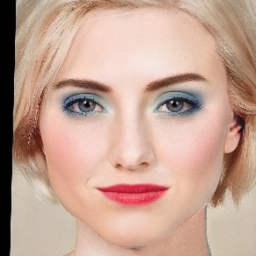}
  \end{subfigure}
  \hfill
  \begin{subfigure}{0.095\textwidth}
    \includegraphics[width=\linewidth]{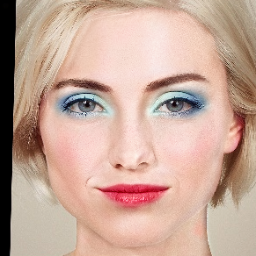}
  \end{subfigure}
  \hfill
  \begin{subfigure}{0.095\textwidth}
    \includegraphics[width=\linewidth]{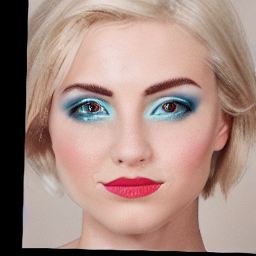}
  \end{subfigure}
  \hfill
  \begin{subfigure}{0.095\textwidth}
    \includegraphics[width=\linewidth]{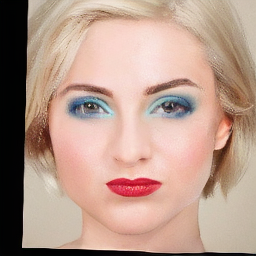}
  \end{subfigure}
  \hfill
  \begin{subfigure}{0.095\textwidth}
    \includegraphics[width=\linewidth]{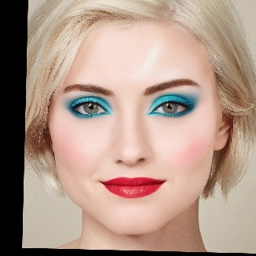}
  \end{subfigure}

  \vspace{0.2mm}

  \begin{subfigure}{0.095\textwidth}
    \captionsetup{font=small}
    \includegraphics[width=\linewidth]{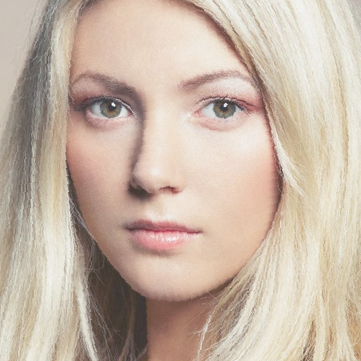}
    \caption*{Source}
  \end{subfigure}
  \hfill
  \begin{subfigure}{0.095\textwidth}
    \captionsetup{font=small}
    \includegraphics[width=\linewidth]{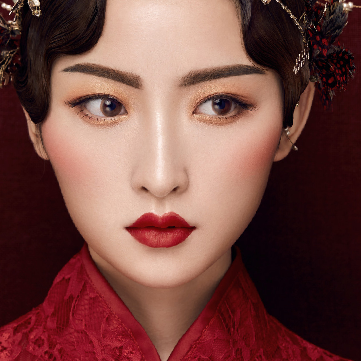}
    \caption*{Reference}
  \end{subfigure}
  \hfill
  \begin{subfigure}{0.095\textwidth}
    \captionsetup{font=small}
    \includegraphics[width=\linewidth]{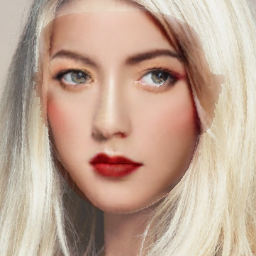}
    \caption*{CPM}
  \end{subfigure}
  \hfill
  \begin{subfigure}{0.095\textwidth}
    \captionsetup{font=small}
    \includegraphics[width=\linewidth]{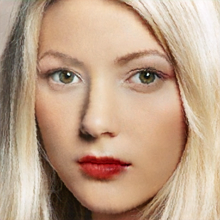}
    \caption*{PSGAN}
  \end{subfigure}
  \hfill
  \begin{subfigure}{0.095\textwidth}
    \captionsetup{font=small}
    \includegraphics[width=\linewidth]{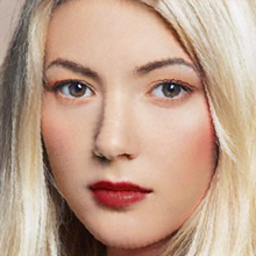}
    \caption*{EleGANt}
  \end{subfigure}
  \hfill
  \begin{subfigure}{0.095\textwidth}
    \captionsetup{font=small}
    \includegraphics[width=\linewidth]{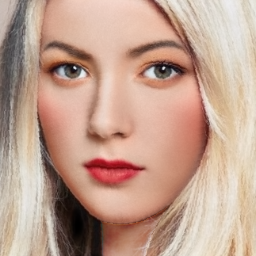}
    \caption*{SSAT}
  \end{subfigure}
  \hfill
  \begin{subfigure}{0.095\textwidth}
    \captionsetup{font=small}
     \includegraphics[width=\linewidth]{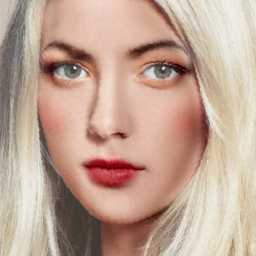}
    \caption*{CSD-MT}
  \end{subfigure}
  \hfill
  \begin{subfigure}{0.095\textwidth}
    \captionsetup{font=small}
     \includegraphics[width=\linewidth]{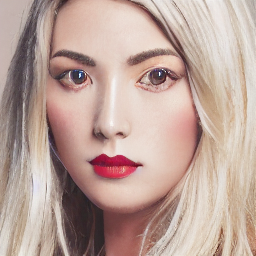}
    \caption*{\scriptsize Stable-Makeup}
  \end{subfigure}
  \hfill
  \begin{subfigure}{0.095\textwidth}
    \captionsetup{font=small}
    \includegraphics[width=\linewidth]{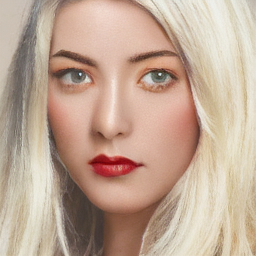}
    \caption*{SHMT}
  \end{subfigure}
  \hfill
  \begin{subfigure}{0.095\textwidth}
    \captionsetup{font=small}
    \includegraphics[width=\linewidth]{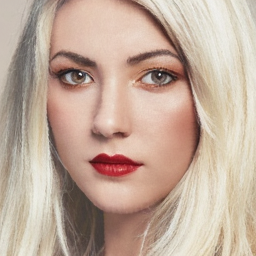}
    \caption*{\scriptsize FLUX-Makeup}
  \end{subfigure}
  
  \caption{Qualitative comparison of the results produced by different methods on simple makeup transfer cases.}
  \label{visul1}
\end{figure*}

\begin{figure*}[!t] 
  \centering

  \begin{subfigure}{0.095\textwidth}
    \includegraphics[width=\linewidth]{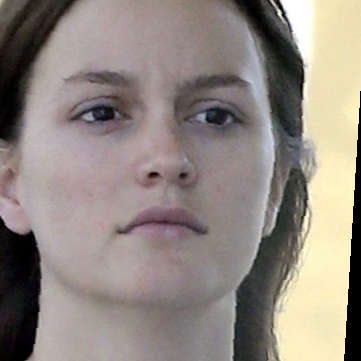}
  \end{subfigure}
  \hfill
  \begin{subfigure}{0.095\textwidth}
    \includegraphics[width=\linewidth]{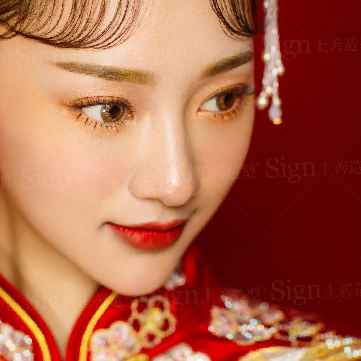}
  \end{subfigure}
  \hfill
  \begin{subfigure}{0.095\textwidth}
    \includegraphics[width=\linewidth]{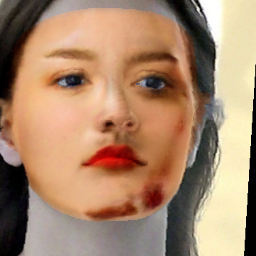}
  \end{subfigure}
  \hfill
  \begin{subfigure}{0.095\textwidth}
    \includegraphics[width=\linewidth]{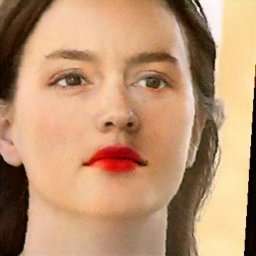}
  \end{subfigure}
  \hfill
  \begin{subfigure}{0.095\textwidth}
    \includegraphics[width=\linewidth]{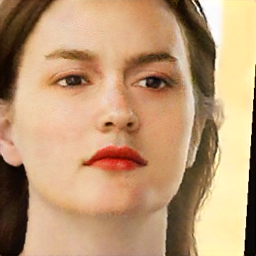}
  \end{subfigure}
  \hfill
  \begin{subfigure}{0.095\textwidth}
    \includegraphics[width=\linewidth]{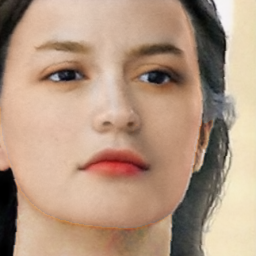}
  \end{subfigure}
  \hfill
  \begin{subfigure}{0.095\textwidth}
    \includegraphics[width=\linewidth]{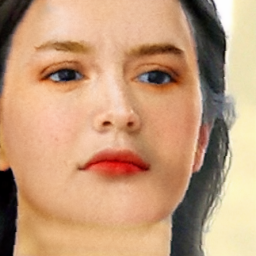}
  \end{subfigure}
  \hfill
  \begin{subfigure}{0.095\textwidth}
    \includegraphics[width=\linewidth]{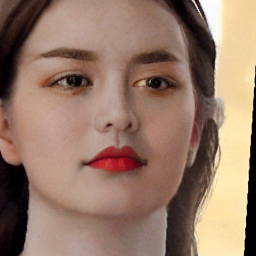}
  \end{subfigure}
  \hfill
  \begin{subfigure}{0.095\textwidth}
    \includegraphics[width=\linewidth]{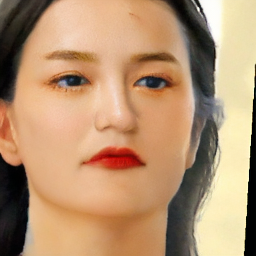}
  \end{subfigure}
  \hfill
  \begin{subfigure}{0.095\textwidth}
    \includegraphics[width=\linewidth]{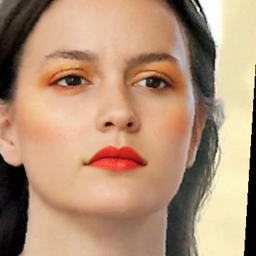}
  \end{subfigure}

  \vspace{0.2mm}

  \begin{subfigure}{0.095\textwidth}
    \includegraphics[width=\linewidth]{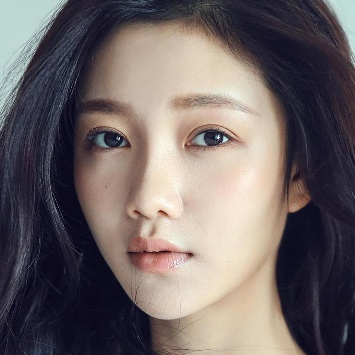}
  \end{subfigure}
  \hfill
  \begin{subfigure}{0.095\textwidth}
    \includegraphics[width=\linewidth]{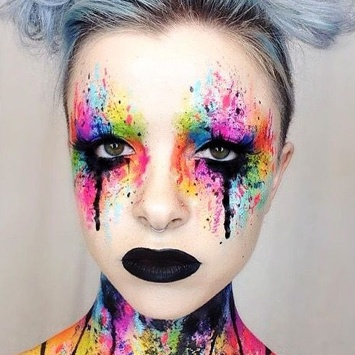}
  \end{subfigure}
  \hfill
  \begin{subfigure}{0.095\textwidth}
    \includegraphics[width=\linewidth]{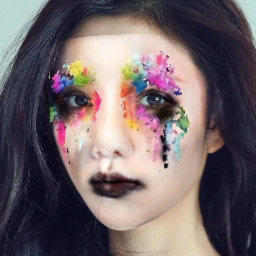}
  \end{subfigure}
  \hfill
  \begin{subfigure}{0.095\textwidth}
    \includegraphics[width=\linewidth]{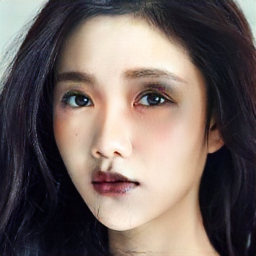}
  \end{subfigure}
  \hfill
  \begin{subfigure}{0.095\textwidth}
    \includegraphics[width=\linewidth]{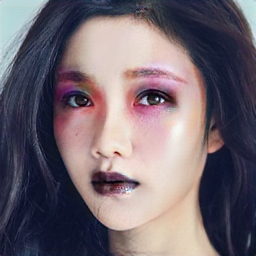}
  \end{subfigure}
  \hfill
  \begin{subfigure}{0.095\textwidth}
    \includegraphics[width=\linewidth]{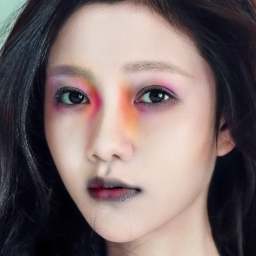}
  \end{subfigure}
  \hfill
  \begin{subfigure}{0.095\textwidth}
    \includegraphics[width=\linewidth]{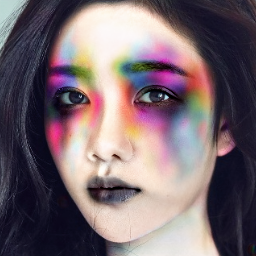}
  \end{subfigure}
  \hfill
  \begin{subfigure}{0.095\textwidth}
    \includegraphics[width=\linewidth]{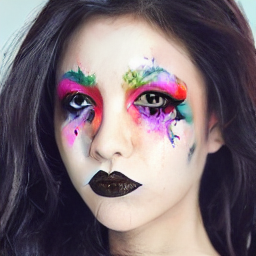}
  \end{subfigure}
  \hfill
  \begin{subfigure}{0.095\textwidth}
    \includegraphics[width=\linewidth]{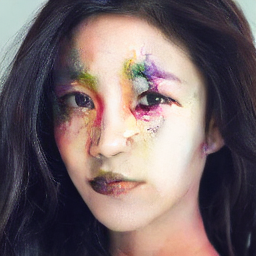}
  \end{subfigure}
  \hfill
  \begin{subfigure}{0.095\textwidth}
    \includegraphics[width=\linewidth]{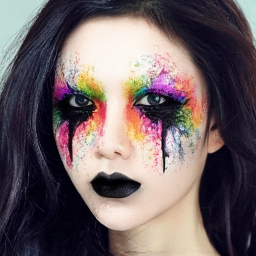}
  \end{subfigure}

  \vspace{0.2mm}
  
  \begin{subfigure}{0.095\textwidth}
    \includegraphics[width=\linewidth]{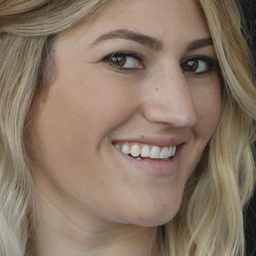}
  \end{subfigure}
  \hfill
  \begin{subfigure}{0.095\textwidth}
    \includegraphics[width=\linewidth]{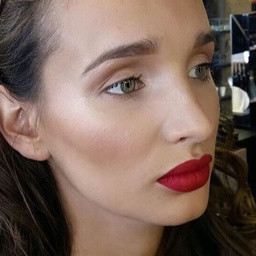}
  \end{subfigure}
  \hfill
  \begin{subfigure}{0.095\textwidth}
    \includegraphics[width=\linewidth]{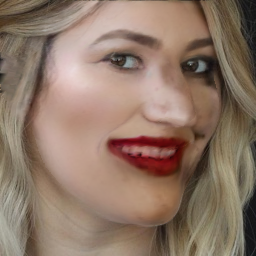}
  \end{subfigure}
  \hfill
  \begin{subfigure}{0.095\textwidth}
    \includegraphics[width=\linewidth]{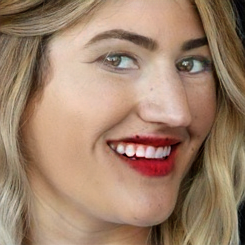}
  \end{subfigure}
  \hfill
  \begin{subfigure}{0.095\textwidth}
    \includegraphics[width=\linewidth]{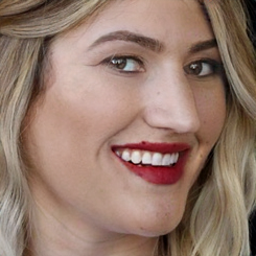}
  \end{subfigure}
  \hfill
  \begin{subfigure}{0.095\textwidth}
    \includegraphics[width=\linewidth]{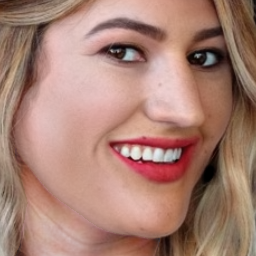}
  \end{subfigure}
  \hfill
  \begin{subfigure}{0.095\textwidth}
    \includegraphics[width=\linewidth]{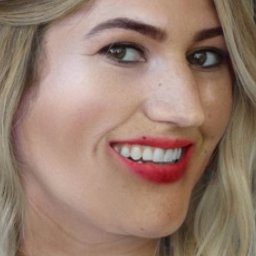}
  \end{subfigure}
  \hfill
  \begin{subfigure}{0.095\textwidth}
    \includegraphics[width=\linewidth]{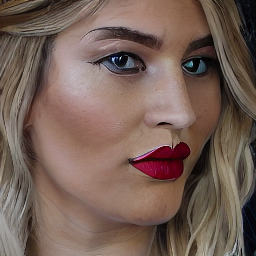}
  \end{subfigure}
  \hfill
  \begin{subfigure}{0.095\textwidth}
    \includegraphics[width=\linewidth]{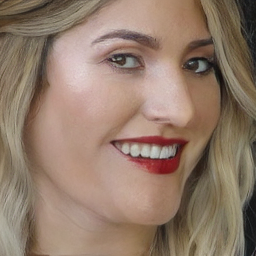}
  \end{subfigure}
  \hfill
  \begin{subfigure}{0.095\textwidth}
    \includegraphics[width=\linewidth]{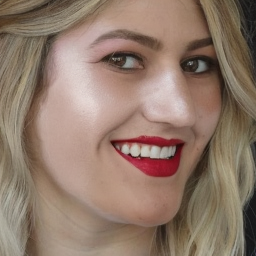}
  \end{subfigure}

  \vspace{0.2mm}

  \begin{subfigure}{0.095\textwidth}
    \captionsetup{font=small}
    \includegraphics[width=\linewidth]{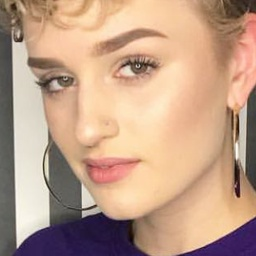}
    \caption*{Source}
  \end{subfigure}
  \hfill
  \begin{subfigure}{0.095\textwidth}
    \captionsetup{font=small}
    \includegraphics[width=\linewidth]{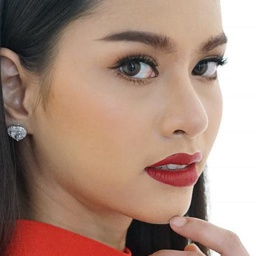}
    \caption*{Reference}
  \end{subfigure}
  \hfill
  \begin{subfigure}{0.095\textwidth}
    \captionsetup{font=small}
    \includegraphics[width=\linewidth]{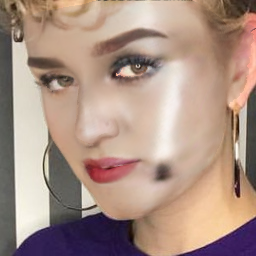}
    \caption*{CPM}
  \end{subfigure}
  \hfill
  \begin{subfigure}{0.095\textwidth}
    \captionsetup{font=small}
    \includegraphics[width=\linewidth]{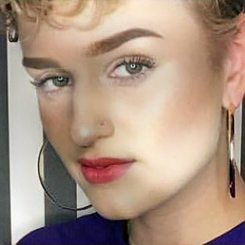}
    \caption*{PSGAN}
  \end{subfigure}
  \hfill
  \begin{subfigure}{0.095\textwidth}
    \captionsetup{font=small}
    \includegraphics[width=\linewidth]{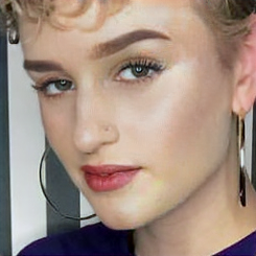}
    \caption*{EleGANt}
  \end{subfigure}
  \hfill
  \begin{subfigure}{0.095\textwidth}
    \captionsetup{font=small}
    \includegraphics[width=\linewidth]{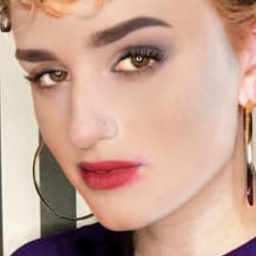}
    \caption*{SSAT}
  \end{subfigure}
  \hfill
  \begin{subfigure}{0.095\textwidth}
    \captionsetup{font=small}
     \includegraphics[width=\linewidth]{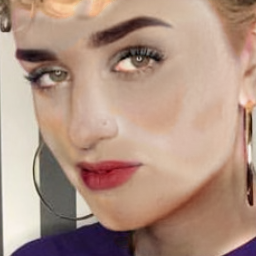}
    \caption*{CSD-MT}
  \end{subfigure}
  \hfill
  \begin{subfigure}{0.095\textwidth}
    \captionsetup{font=small}
     \includegraphics[width=\linewidth]{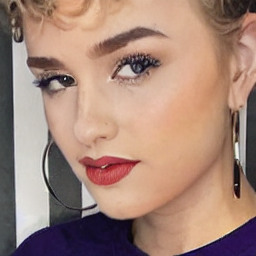}
    \caption*{\scriptsize Stable-Makeup}
  \end{subfigure}
  \hfill
  \begin{subfigure}{0.095\textwidth}
    \captionsetup{font=small}
    \includegraphics[width=\linewidth]{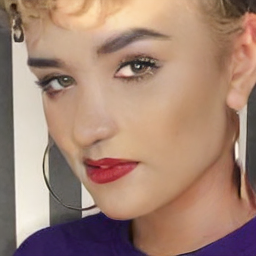}
    \caption*{SHMT}
  \end{subfigure}
  \hfill
  \begin{subfigure}{0.095\textwidth}
    \captionsetup{font=small}
    \includegraphics[width=\linewidth]{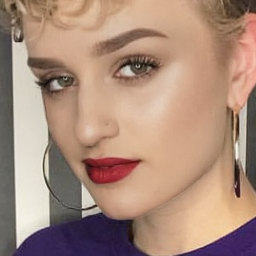}
    \caption*{\scriptsize FLUX-Makeup}
  \end{subfigure}
  
  \caption{Qualitative comparison of the results produced by different methods on challenging makeup transfer cases.}
  \label{visul2}
\end{figure*}

\section{Experiment}
\subsection{Experimental settings}
\paragraph{Datasets.}
We employ the previously described data generation pipeline to construct our training dataset. Starting from the FFHQ dataset, which contains approximately 70,000 face images, we randomly assign five makeup-related adjectives from our curated prompt library to each image. Using the FLUX-Kontext model with the template “\{\} makeup.”, we generate stylized outputs, resulting in a total of 350,000 synthesized makeup images. Subsequently, we apply our proposed three-stage filtering algorithm to remove misaligned, low-quality, or ineffective makeup pairs. After filtering, approximately 56,900 high-quality image pairs remain for training. For clarity, we refer to this final filtered dataset as HQMT (High-Quality Makeup Transfer).

To evaluate the effectiveness and generalizability of our method, we conduct experiments on three widely used makeup transfer benchmarks: MT \cite{li2018beautygan}, Wild‑MT \cite{jiang2020psgan}, and LADN \cite{gu2019ladn}. The MT dataset consists of 3,834 face images in total, including 1,115 non-makeup and 2,719 makeup images. Wild‑MT offers a diverse set of makeup images with varied poses, expressions, and complex backgrounds. The LADN dataset contains unpaired face images before and after makeup, specifically collected to represent complex and detailed cosmetic styles.

\paragraph{Evaluation Metrics.}
To comprehensively and objectively evaluate the performance of different methods, we adopt three quantitative metrics: CLIP-I \cite{radford2021learning}, SSIM \cite{wang2004image}, and L2-M \cite{zhang2025stablemakeup}.
CLIP-I measures the effectiveness of makeup transfer by computing the cosine similarity between the CLIP-encoded features of the reference and generated images.
SSIM evaluates identity preservation by measuring the structural similarity between the source and generated images, focusing on luminance, contrast, and structural information to quantify how well facial identity is maintained.
L2-M assesses background consistency by applying a face parsing algorithm to isolate background regions in both source and generated images, then computing the mean squared error (MSE) between these regions.

\paragraph{Implementation Details.}
We adopt FLUX-Kontext as the base architecture and fine-tune it using our proposed training dataset. The training is conducted on 8 NVIDIA H100 GPUs with a batch size of 4 per GPU. We use the Prodigy optimizer \cite{mishchenko2023prodigy} with a learning rate of 1 and train the model for 100,000 iterations. During training, the text prompt is fixed to “Makeup.” and the LoRA rank in RefLoRAInjector is set to 256. To avoid background interference, we use a face parsing algorithm to extract only the facial region as the reference input.
At inference, we employ 28 sampling steps and a guidance scale of 2.5.

\subsection{Experimental Results}
In this work, we conduct a comprehensive comparison between our method and seven state-of-the-art makeup transfer approaches, namely CPM \cite{nguyen2021lipstick}, PSGAN \cite{jiang2020psgan}, EleGANt \cite{yang2022elegant}, SSAT \cite{sun2022ssat}, CSD-MT \cite{sun2024content}, Stable-Makeup \cite{zhang2025stablemakeup}, and SHMT \cite{sun2024shmt}. Among these methods, Stable-Makeup and SHMT adopt diffusion-based architectures, whereas the others are built upon GAN frameworks. For fairness and reproducibility, all comparison results are obtained using the publicly available implementations and pretrained models provided by the respective authors.

\paragraph{Quantitative Comparison.} 
To quantitatively evaluate the makeup transfer capability of our model, we randomly selected 1,000 source–reference image pairs from each of the three datasets to generate makeup transfer results. These results were assessed using three metrics: CLIP-I, SSIM, and L2-M, as shown in Table \ref{compare}. CLIP-I primarily reflects the fidelity of makeup transferred from the reference image. The results show that diffusion-based methods generally achieve higher transfer fidelity than GAN-based methods, and our approach, FLUX-Makeup, further exceeds the two diffusion-based baselines, Stable-Makeup and SHMT, in this aspect.
For consistency evaluation, reflected by SSIM and L2-M, FLUX-Makeup significantly outperforms all other compared methods, demonstrating its strong ability to maintain consistency during makeup transfer. Among the other methods, CPM achieves relatively good consistency but still exhibits a noticeable gap compared to our approach.

\paragraph{Qualitative Comparison.}
To comprehensively evaluate the performance of our model, we conducted qualitative comparisons on both simple and challenging makeup transfer cases, with results shown in Fig. \ref{visul1} and Fig. \ref{visul2}, respectively. In the simpler cases, although CPM successfully completes the transfer, it often introduces noticeable artifacts. PSGAN, EleGANt, SSAT, and CSD-MT demonstrate limited effectiveness in transferring makeup. Stable-Makeup and SHMT achieve better results; however, they frequently cause unintended changes in facial features, particularly around the mouth area.

The challenging cases presented in Fig. \ref{visul2}, involving complex makeup styles and diverse poses, offer a more stringent evaluation. Under these conditions, GAN-based methods suffer a significant drop in performance, generating more artifacts and exposing their lack of robustness in handling complex scenarios. Diffusion-based methods such as Stable-Makeup and SHMT produce more coherent results but still fall short of our approach in terms of makeup fidelity and consistency.
This qualitative analysis clearly demonstrates that FLUX-Makeup excels at maintaining high-quality makeup transfer while preserving identity and background consistency, showing strong robustness across various makeup transfer scenarios.

\subsection{Ablation Study}
\begin{figure}[!t] 
  \centering
  
  \begin{subfigure}{0.091\textwidth}
    \includegraphics[width=\linewidth]{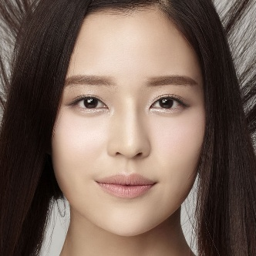}
  \end{subfigure}
  \hfill
  \begin{subfigure}{0.091\textwidth}
    \includegraphics[width=\linewidth]{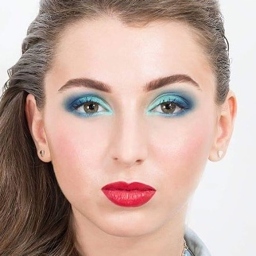}
  \end{subfigure}
  \hfill
  \begin{subfigure}{0.091\textwidth}
    \includegraphics[width=\linewidth]{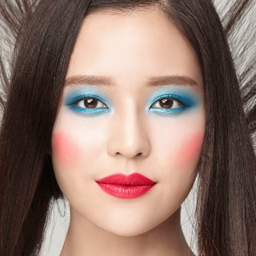}
  \end{subfigure}
  \hfill
  \begin{subfigure}{0.091\textwidth}
    \includegraphics[width=\linewidth]{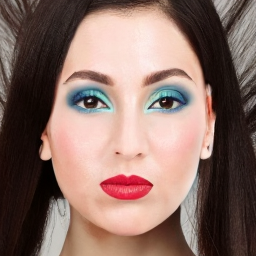}
  \end{subfigure}
  \hfill
  \begin{subfigure}{0.091\textwidth}
    \includegraphics[width=\linewidth]{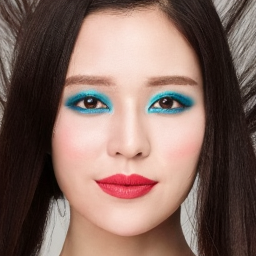}
  \end{subfigure}

  \vspace{0.2mm}

  \begin{subfigure}{0.091\textwidth}
    \captionsetup{font=small}
    \includegraphics[width=\linewidth]{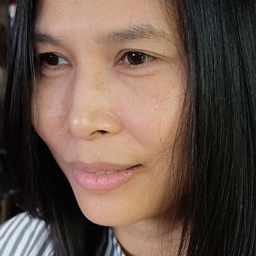}
    \caption*{Source}
  \end{subfigure}
  \hfill
  \begin{subfigure}{0.091\textwidth}
    \captionsetup{font=small}
    \includegraphics[width=\linewidth]{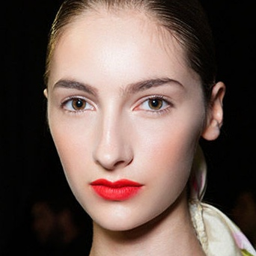}
    \caption*{Reference}
  \end{subfigure}
  \hfill
  \begin{subfigure}{0.091\textwidth}
    \captionsetup{font=small}
     \includegraphics[width=\linewidth]{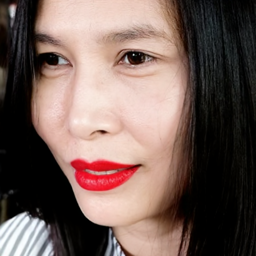}
    \caption*{\scriptsize FLUX}
  \end{subfigure}
  \hfill
  \begin{subfigure}{0.091\textwidth}
    \captionsetup{font=small}
    \includegraphics[width=\linewidth]{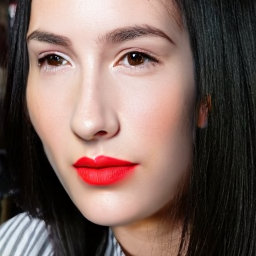}
    \caption*{\scriptsize FLUX-Kontext}
  \end{subfigure}
  \hfill
  \begin{subfigure}{0.091\textwidth}
    \captionsetup{font=small}
    \includegraphics[width=\linewidth]{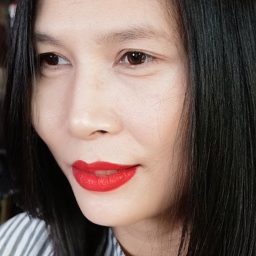}
    \caption*{\scriptsize FLUX-Makeup}
  \end{subfigure}
  
  \caption{Visual comparison of results from the ablation study on different framework designs.}
  \label{abv}
\end{figure}

\paragraph{Framework Ablation.}
As illustrated in Fig. \ref{fig1}, we initially investigated two alternative frameworks for the makeup transfer task: the mainstream approach of fine-tuning FLUX with LoRA paradigm \cite{zhang2025easycontrol}, and a straightforward variant based on FLUX-Kontext with LoRA fine-tuning. The ablation results are summarized in Table \ref{ab1} and Fig. \ref{abv}.
The FLUX-based LoRA fine-tuning approach exhibits lower transfer accuracy and weaker consistency preservation. On the other hand, although the FLUX-Kontext–based LoRA fine-tuning achieves a higher CLIP-I score than our method, visual inspection reveals that this stems mainly from excessive enforcement of consistency, which effectively pastes the reference face directly onto the source image.
In contrast, our framework achieves a better balance between transfer fidelity and consistency, resulting in high-quality and faithful makeup transfer.

\begin{table}[htbp] 
    \centering
    \begin{tabular}{lccc}
        \toprule
        Method  & CLIP-I & SSIM & L2-M  \\
        \midrule
        FLUX + LoRA            &0.708 &0.806   &14.23  \\
        FLUX-Kontext + LoRA    &0.784 &0.787   &9.09   \\
        FLUX-Makeup            &0.731 &0.862   &5.20   \\
        \bottomrule
    \end{tabular}
    \caption{Ablation study on different framework designs for the makeup transfer task.}
    \label{ab1}  
\end{table}

\paragraph{Training Data Ablation.}
To further validate the effectiveness of our data generation algorithm, we first evaluated the pass rate (PR) of different datasets. Specifically, 500 images were randomly sampled from the Stable-Makeup dataset, the unfiltered HQMT dataset, and our HQMT dataset for manual inspection, where samples with background inconsistencies, facial identity alterations, or missing makeup were judged as unqualified. As shown in Table \ref{ab2}, our HQMT dataset demonstrates a substantially higher PR, outperforming the other two datasets by a large margin.
We further trained and evaluated our model on these three datasets, with performance metrics measured on the LADN benchmark. The results clearly indicate that our HQMT dataset delivers more accurate and reliable supervision during training, significantly improving makeup transfer performance while preserving both facial features and background consistency.

\begin{table}[!t] 
    \centering
    \begin{tabular}{lcccc}
        \toprule
        Dataset  & PR (\%) & CLIP-I & SSIM & L2-M \\
        \midrule
        Stable-Makeup      &6.8  &0.521  &0.496 &45.62 \\
        Unfiltered HQMT    &15.4 &0.615  &0.736 &17.93 \\
        HQMT               &96.2 &0.731  &0.862 &5.20  \\
        \bottomrule
    \end{tabular}
    \caption{Ablation study on the impact of different training datasets on model performance.}
    \label{ab2}  
\end{table}

\section{Conclusion}
In this paper, we propose FLUX-Makeup, a high-fidelity makeup transfer method built upon the DiT architecture. Unlike many existing approaches, it operates without relying on any face control modules, thereby avoiding unnecessary error accumulation and preserving identity more accurately. To further enhance the model’s capability, we introduce RefLoRAInjector, a lightweight makeup feature embedding module that decouples the reference feature extraction path from the main backbone, enabling more precise makeup information capture while significantly reducing computational overhead.
Moreover, we develop a robust and scalable data generation pipeline that produces high-quality paired makeup datasets. This pipeline not only ensures accurate alignment and realistic makeup effects but also yields data quality far superior to existing publicly available datasets. Extensive experiments across multiple benchmarks confirm that our method achieves outstanding results in both makeup fidelity and consistency, consistently surpassing current state-of-the-art approaches.

\bibliography{aaai2026}

\end{document}